\documentclass{article}

\usepackage[english]{babel}
\usepackage[utf8]{inputenc}
\usepackage[T1]{fontenc}

\usepackage{amsmath}
\usepackage{amssymb}
\usepackage{amsthm}
\usepackage{amsfonts}
\usepackage{graphicx}
\usepackage{hyperref}
\usepackage[numbers,square]{natbib}
\usepackage{bbm}
\usepackage{amsthm}
\usepackage{authblk}
\usepackage{times}
\usepackage{authblk}
\usepackage{algorithmic}
\usepackage[squaren]{SIunits}

\newcommand{\BEAS}{\begin{eqnarray*}}
\newcommand{\EEAS}{\end{eqnarray*}}
\newcommand{\BEA}{\begin{eqnarray}}
\newcommand{\EEA}{\end{eqnarray}}
\newcommand{\BEQ}{\begin{equation}}
\newcommand{\EEQ}{\end{equation}}
\newcommand{\BIT}{\begin{itemize}}
\newcommand{\EIT}{\end{itemize}}
\newcommand{\BNUM}{\begin{enumerate}}
\newcommand{\ENUM}{\end{enumerate}}
\newcommand{\BA}{\begin{array}}
\newcommand{\EA}{\end{array}}

\DeclareMathOperator*{\argmax}{argmax}
\DeclareMathOperator*{\argmin}{argmin}

\DeclareMathOperator{\Tr}{Tr}
\DeclareMathOperator{\tr}{Tr}

\DeclareMathOperator{\deltamax}{\delta_{\max}}
\DeclareMathOperator{\deltaabs}{\delta_{\mathrm{abs}}}

\newcommand{\R}{\mathbb{R}}

\newcommand*\samethanks[1][\value{footnote}]{\footnotemark[#1]}

\title{Metric Learning for Temporal Sequence Alignment}
\author[1,2]{Damien Garreau \thanks{Both authors contributed equally.} \\ \texttt{damien.garreau@ens.fr}}

\author[1,2]{\\ R\'emi Lajugie \samethanks \\ \texttt{remi.lajugie@inria.fr}}
 \author[1,2]{\\ Sylvain Arlot \\ \texttt{sylvain.arlot@ens.fr}}
\author[1,2]{\\ Francis Bach \\  \texttt{francis.bach@inria.fr}}
\affil[1]{Département d'Informatique de l'Ecole Normale Supérieure}
\affil[2]{SIERRA project team, Inria Paris Rocquencourt}
\begin{document}
\maketitle
\begin{abstract}

In this paper, we propose to learn a Mahalanobis distance to perform alignment of multivariate time series. The learning examples for this task are time series for which the true alignment is known. We cast the alignment problem as a structured prediction task, and propose realistic losses between alignments for which the optimization is tractable. We provide experiments on real data in the audio to audio context, where we show that the learning of a similarity measure leads to improvements in the performance of the alignment task. We also propose to use this metric learning framework to perform feature selection and, from basic audio features, build a combination of these with better performance for the alignment.

\end{abstract}

\section{Introduction}

The problem of aligning temporal sequences is ubiquitous in applications ranging from bioinformatics \citep{cuturi2007kernel,aach2001aligning,thompson1999balibase} to audio processing \citep{cont2007evaluation,dixon2005match}. The idea is to align two similar time series that have the same global structure but local temporal differences. Most alignments algorithms rely on similarity measures, and having a good metric is crucial, especially in the high-dimensional setting where some features of the signals can be irrelevant to the alignment task. The goal of this paper is to show how to learn this similarity measure from annotated examples in order to improve the precision of the alignments.

For example, in the context of music information retrieval, alignment is used in two different cases: (1) audio-to-audio alignment and (2) audio-to-score alignment. In the first case, the goal is to match two audio interpretations of the same piece that are potentially different in rythm, whereas audio-to-score alignment focuses on matching an audio signal to a symbolic representation of the score. In the second case, there are some attempts to learn from annotated data a measure for performing the alignment. \citet{joder2013learning} propose to fit a generative model in that context, and \citet{keshet2007large} learn this measure in a discriminative setting. 
%

%
%
Similarly to \citet{keshet2007large}, we use a discriminative loss to learn the measure, but our work focuses on audio-to-audio alignment. In that context, the set of authorized alignments is much larger, and we explicitly cast the problem as a structured prediction task, that we solve using off-the-shelf stochastic optimization techniques \citep{lacoste2013block} but with proper and significant adjustements, in particular in terms of losses.
The need for metric learning goes far beyond unsupervised partitioning problems. \citet{weinberger2009distance} proposed a large-margin framework for learning a metric in nearest-neighbour algorithms based on sets of must-link/must-not-link constraints. \citet{lajugie2014large} proposed to use a large margin framework to learn a Mahalanobis metric in the context of partitioning problems. Since structured SVM have been proposed by~\citet{tsochantaridis2005large, koller2003max}, they have successfully been used to solve many learning problems, for instance to learn weights for graph matching~\citep{caetano2009learning} or a metric for ranking tasks~\citep{mcfee2010metric}. 
They have also been used to learn graph structures using graph cuts~\citep{szummer2008learning}. 

\paragraph{Contributions.}

We make the following five contributions: (1) we cast the learning of a Mahalanobis metric in the context of alignment as a structured prediction problem, (2) we show that on real musical datasets this metric improves the performance of alignment algorithms using high-level features, (3) we propose to use the metric learning framework to learn combinations of basic audio features and get good alignment performances, (4) we show experimentally that the standard Hamming loss, although tractable computationnally does not permit to learn a relevant similarity measure in some real world settings, (5) we propose a new loss, closer to the true evaluation loss for alignments, leading to a tractable learning task, and derive an efficient Frank-Wolfe based algorithm to deal with this new loss.

\section{Matricial formulation of alignment problems}

\subsection{Notations}
\label{sec:notations}
In this paper, we consider the alignment problem between two multivariate time series sharing the same dimension $p$, but possibly of different lengths $T_A$ and $T_B$, namely $A \in \mathbb{R}^{T_A \times p}$ and $B \in \mathbb{R}^{T_B \times p}$. We refer to the rows of $A$ as $a_1,\ldots, a_{T_A} \in \mathbb{R}^p$ and those of $B$ as $b_1, \ldots, b_{T_B}\in \mathbb{R}^p$. From now on, we denote by $X$ the pair of signals $(A,B)$.

Let $C(X)\in\R^{T_A\times T_B}$ be an arbitrary pairwise \emph{affinity matrix}  associated to the pair $X$, that is, $C(X)_{i,j}$ encodes the affinity between $a_i$ and $b_j$. 
Note our framework can be extended to the case where $A$ and $B$ are multivariate signals of different dimensions, as long as $C(X)$ is well-defined. The goal of the alignment task is to find two non-decreasing sequences of indices $\alpha$ and $\beta$ of same length $u\geq\max(T_A,T_B)$ and to match each time index $\alpha(i)$ in time series $A$ to the time index $\beta(i)$ in the time series $B$, in such a way that $\sum^u_{i=1} C(X)_{\alpha(i),\beta(i)}$ is maximal, and that $(\alpha,\beta)$ satisfies:

\BEQ
	\left\{
		\begin{aligned}
			\alpha(1) =&\;\beta(1) = 1 &&\text{(matching beginning)}\\
			\alpha(u) = T_A,&\; \beta(u) = T_B &&\text{(matching ending)}\\
			\forall i, \, (\alpha(i+1), \beta(i+1)) - (\alpha(i),\beta(i)) &\in \left\{(1,0),(0,1),(1,1)\right\} &&\text{(three type of moves)}
		\end{aligned}
	\right.
\label{eq:constraint}
\EEQ

\begin{figure}
\caption{\label{fig:losses} Example of two valid alignments encoded by matrices $Y^1$ and $Y^2$. Red upper triangles show the $(i,j)$ such that $Y^1_{i,j}=1$, and the blue lower ones show the $(i,j)$ such that $Y^2_{i,j}=1$. The grey zone corresponds to the area loss $\deltaabs$ between $Y^1$ and $Y^2$, whereas the $\deltamax$ loss corresponds to the maximum of the $\delta_{t}$, $t\in\{1,\dots,T_1\}$.}
\begin{center}
\includegraphics[scale=0.33]{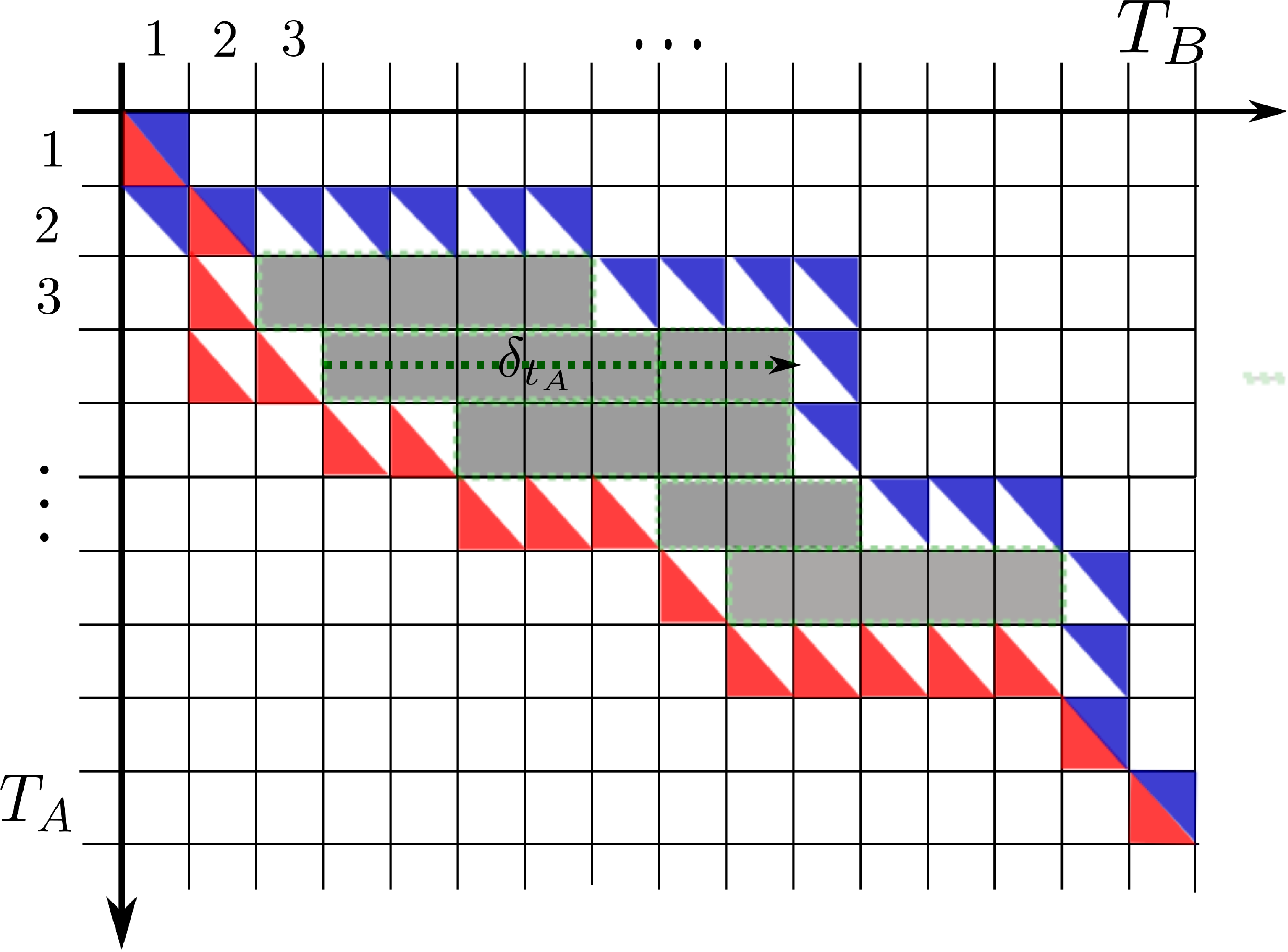}
\end{center}
\end{figure}

For a given $(\alpha,\beta)$, we define the binary matrix $Y \in \{0, 1\}^{T_A \times T_B}$ such that $Y_{\alpha(i),\beta(i)} = 1$ for every $i\in\{1,\dots,u\}$ and $0$ otherwise. We denote by $\mathcal{Y}(X)$ the set of such matrices, which is uniquely determined by $T_A$ and $T_B$. An example is given in Fig.~\ref{fig:losses}. A vertical move in the $Y$ matrix means that the signal $B$ is waiting for $A$ and an horizontal one that $A$ is waiting for $B$. In this sense we can say that the time reference is ``warped''.

When $C(X)$ is known, the alignment task can be cast as the following linear program (LP) over the set $\mathcal{Y}(X)$:
\BEQ \label{eq:decoding} \max_{Y \in \mathcal{Y}(X)} \Tr(C(X)^\top Y).
\EEQ
Our goal is to learn how to form the affinity matrix: once we have learned $C(X)$, the alignment is obtained from Eq.~\eqref{eq:decoding}. The optimization problem in Eq.~\eqref{eq:decoding} will be referred to as the \emph{decoding} of our model.

\paragraph{Dynamic time warping.}

Given the affinity matrix $C(X)$ associated with the pair of signals $X=(A,B)$, finding the alignment that solves the LP of Eq.~\eqref{eq:decoding} can be done efficiently in $O(T_AT_B)$ using a dynamic programming algorithm. It is often referred to as dynamic time warping \citep{cuturi2007kernel,muller2007information}. This algorithm is described in Alg.~1 of the supplementary material. Various additional constraints may be used in the dynamic time warping algorithm \citep{muller2007information}, which we could easily add.

The cardinality of the set $\mathcal{Y}(X)$ is huge: it corresponds to the number of paths on a rectangular grid from the southwest $(1,1)$ to the northeast corner $(T_A,T_B)$ with only vertical, horizontal and diagonal moves allowed. This is the definition of the Delannoy numbers~\citep{banderier2005delannoy}. As noted in \cite{torres2003exact}, when $t=T_A=T_B$ is big, one can show that
$ 
\#\mathcal{Y}_{t,s}\sim \frac{(3+2\sqrt{2})^t}{\sqrt{\pi t}\sqrt{3\sqrt{2} - 4}}
.$

\subsection{The Mahalanobis metric}

In many applications, for a pair $X=(A,B)$, the affinity matrix is computed by $C(A,B)_{i,j} = -\|a_{i,k} - b_{j,k}\|^2$. In this paper we propose to learn the metric to compare $a_i$ and $b_j$ instead of using the plain Euclidean metric. That is, $C(X)$ is parametrized by a matrix $W \in \mathcal{W} \subset \mathbb{R}^{p \times p}$, where $\mathcal{W}\subset\R^{p\times p}$ is the set of semi-definite positive matrices, and we use the corresponding Mahalanobis metric to compute the pairwise affinity between $a_i$ and $b_j$: $C(X;W)_{i,j} = -(a_i - b_j)^\top W(a_i - b_j)$.

Note that the decoding of Eq.~\eqref{eq:decoding} is the maximization of a linear function in the parameter $W$:

\BEQ\label{eq:decodingWithw}
\max_{Y \in \mathcal{Y}(X)}\Tr(C(X;W)^\top Y) \Leftrightarrow \max_{Y \in \mathcal{Y}(X)}\Tr(W^\top \phi(X,Y)),
\EEQ

if we define the joint feature map $\phi(X,Y)=-\sum_{i,j} Y_{i,j}(a_i - b_j)(a_i - b_j)^\top\in\R^{p\times p}$. 

\section{Learning the metric}

From now on, we assume that we are given $n$ pairs of training instances\footnote{We will see that it is necessary to have fully labelled instances, which means that for each pair $X^i$ we need an \emph{exact} alignment $Y^i$ between $A^i$ and $B^i$. Partial alignment might be dealt with by alternating between metric learning and constrained alignment.} $(X^i,Y^i)=((A^i, B^i), Y^i) \in \mathbb{R}^{T^i_A \times p}\times\mathbb{R}^{T^i_B \times p} \times \mathbb{R}^{T^i_A \times T_B^i}$, $i = 1, \ldots, n$. 
Our goal is to find a matrix $W$ such that the predicted alignments are close to the groundtruth on these examples, as well as on unseen examples. We first define a \emph{loss} between alignments, in order to quantify this proximity between alignments.

\subsection{Losses between alignments}

In our framework, the alignments are encoded by matrices in $\mathcal{Y}(X)$, thus we are interested in functions $\ell:\mathcal{Y}(X) \times \mathcal{Y}(X) \to \R_+$.  Let us define the Frobenius norm by $\|M\|_F^2 = \sum_{i,j}M_{i,j}^2$.

\paragraph{Hamming loss.}

A simple loss between matrices is the Frobenius norm of their difference, which turns out to be the unnormalized Hamming loss~\citep{hamming1950error} for $0/1$ valued matrices.
For two matrices $Y_1, Y_2 \in \mathcal{Y}(X)$, it is defined as:
\begin{align}
\label{eq:Hammingloss}
\ell_H(Y_1, Y_2) &= \|Y_1 - Y_2\|^2_F\\
\notag 
				&= \Tr(Y_1^\top Y_1) + \Tr(Y_2^\top Y_2) - 2 \Tr(Y_1^\top Y_2)\\
\notag 
			   &= \Tr(Y_1\mathbf{1}_{T_B}\mathbf{1}_{T_A}^\top) + \Tr(Y_2\mathbf{1}_{T_B}\mathbf{1}_{T_A}^\top) - 2 \Tr(Y_1^\top Y_2),
\end{align}
where $\mathbf{1}_T$ is the vector of $\R^T$ with all coordinates equal to $1$. The last line of Eq.~\eqref{eq:Hammingloss} comes from the fact that the $Y_i$ have $0-1$ values; that makes the Hamming loss linear in $Y_1$ and $Y_2$. This loss is often used in other structured prediction tasks~\citep{lacoste2013block}. In the audio-to-score setting, \citet{keshet2007large} use a modified version of this loss, which is the average number of times the difference between the two alignments is greater than a fixed threshold.

\begin{figure}
\begin{center}
\includegraphics[scale=.33]{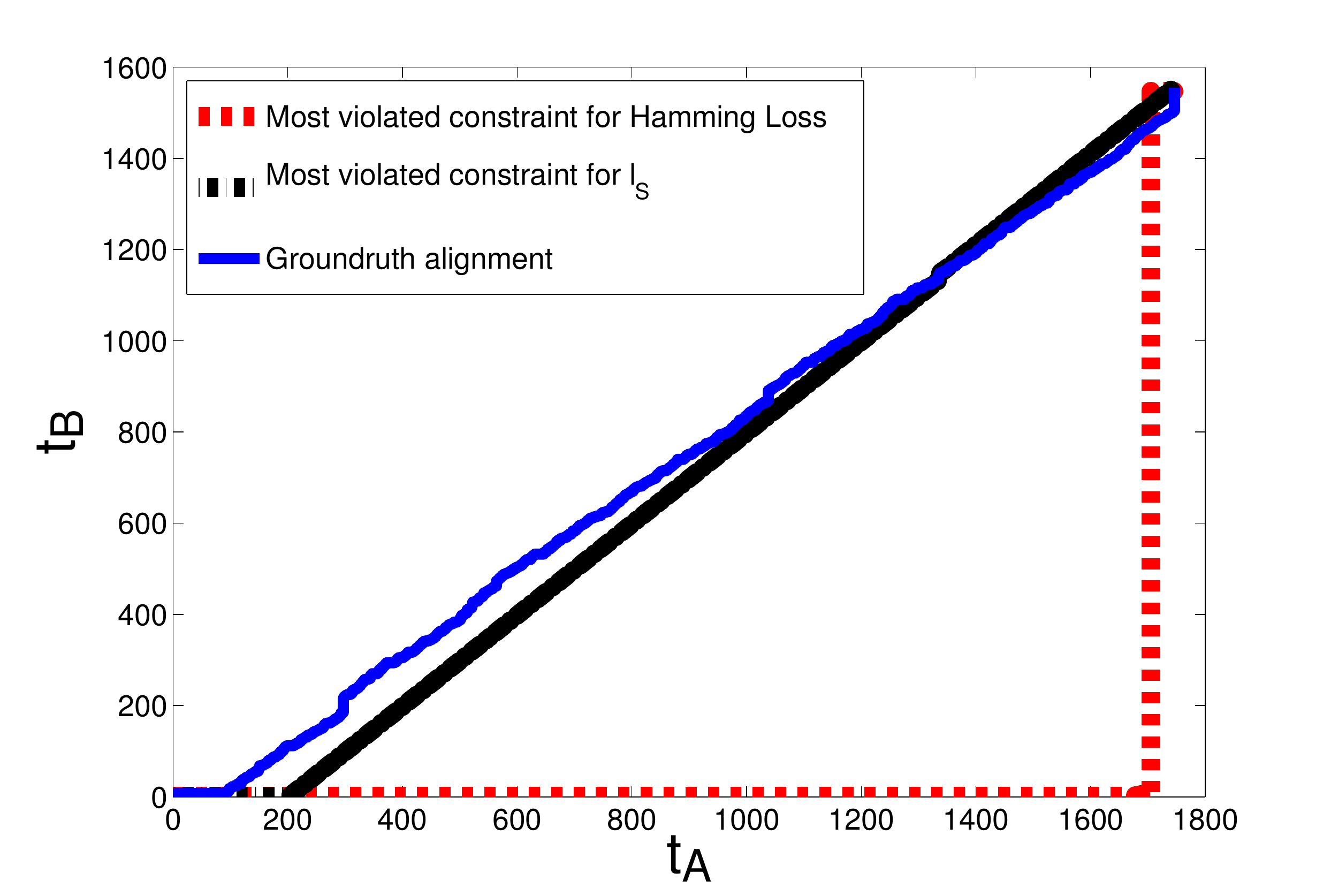}
\caption{\label{fig:hamming_fail} On the real world Bach chorales dataset, we have represented the most violated constrained at the end of learning, when the training loss is the Hamming one or the symmetrized area loss. Note also that, in terms of Hamming loss the most violated contraint for $\ell_S$ and the Hamming one are the same.}
\end{center}
\end{figure}

This loss is easy to optimize since it is linear in our parametrization of the alignement problem, but not optimal for audio-to-audio alignment. 
Indeed, a major drawback of the Hamming loss is, for alignments of fixed length, it depends only on the number of ``crossings'' between alignment paths: one can easily find $Y_1, Y_2, Y_3$ such that $\ell_H(Y_2,Y_1)=\ell_H(Y_3,Y_1)$ but $Y_2$ is intuitively much closer to $Y_1$ than $Y_3$ (see Fig.~\ref{fig:hamming_fail}). 
It is important to notice this is often the case as the length of the signals grows.


\paragraph{Area loss.}
A more natural loss can be computed as the mean distance beween the paths depicted by two matrices $Y^1, Y^2 \in \mathcal{Y}(X)$. This loss is represented by the grey zone on Fig.~\ref{fig:losses} and corresponds to the area between the paths of two matrices $Y$. 

Formally, as in Fig.~\ref{fig:losses}, for each $t\in\{1,\dots,T_B\}$ we define $\delta_{t}$ as the minimum between $|\min \{k, Y_{k, t}^1 = 1\} - \max \{k, Y_{k, t}^2 = 1\}|$ and $|\max \{k, Y_{k, t}^1 = 1\} - \min \{k, Y_{k, t}^2 = 1\}|$. 
Then the area loss is the mean of the $\delta_{t}$. In the audio literature \citep{kirchhoff2011evaluation}, this loss is sometimes called the ``mean absolute deviation'' loss and is noted $\deltaabs(Y^1, Y^2)$.

Unfortunately, in the general case of alignment problem $\deltaabs$ is not linear in the matrices $Y$. But in the context of alignment of sequences of two different nature, one of the signal is a reference and thus the index sequence $\alpha$ defined in Eq.~\eqref{eq:constraint} is increasing, e.g. for the audio to partition alignment problem \citep{joder2013learning}. This loss is then linear in each of its arguments. More precisely, if we introduce the matrices $L\in\R^{T_A\times T_A}$ which is lower triangular with ones, we can write the loss as 
\begin{align}
\label{eq:arealoss}
\ell_O &= \|L(Y_1 - Y_2)\|_F^2 \\
\notag 
&= \Tr(LY_1\mathbf{1}_{T_B}\mathbf{1}_{T_A}^\top) + \Tr(LY_2 \mathbf{1}_{T_B}\mathbf{1}_{T_A}^\top)- 2 \Tr(L Y_1 Y_2^\top L^\top).
\end{align}
We now prove that this loss corresponds to the area loss in this special case. Let $Y$ be an alignment matrix and $L\in\R^{T_A\times T_A}$ be the matrix such that $L_{r,s} = 1$ if and only if $r\geq s$. 
Then it is easy see that $(LY)_{i,j} = \sum_k L_{i,k}Y_{k,j} = \sum_{k=0}^iY_{k,j}$. 
If $Y$ does not have vertical moves, i.e. for each $j$ there is an unique $k_j$ such that $Y_{k,j} = 1$, we have that $(LY)_{i,j}=1$ if and only if $i\geq k_j$.
So $\sum_{i,j} (LY)_{i,j} = \#\{(i,j), i\geq k_j\}$, which is exactly the area under the curve determined by the path of $Y$.
 

In all our experiments, we use $\deltaabs$ for evaluation but not for training.

\paragraph{Approximation of the area loss: the symmetrized area loss.}

In many real world applications \cite{kirchhoff2011evaluation}, the best loss to assess the quality of an alignment is the area loss. 
As shown by our experiments, if the Hamming loss is sufficient in some simple situations and allows to learn a metric that leads to good alignment performance in terms of area loss, on more challenging datasets it does not work at all (see Sec.~\ref{sec:experiments}). 
This is due to the fact that two alignments that are very close in terms of area loss can suffer a big Hamming loss. 
In Fig.~\ref{fig:hamming_fail}, we provide examples where the Hamming loss is not sufficient to assess performance.
Thus it is natural to extend the formulation of Eq.~\eqref{eq:arealoss} to matrices in $\mathcal{Y}(X)$.
We first start by symmetrizing the formulation of Eq.~\eqref{eq:arealoss} to overcome problems of overpenalization of vertical vs. horizontal moves. Let $L_1\in\R^{T_B\times T_B}$ be the matrix such that $L_{r,s} = 1$ if and only if $r\geq s$. We define, for any binary matrices $Y^1$ and $Y^2$, 
\begin{align}
\label{eq:symloss}
\ell_S(Y_1,Y_2) &= \frac{1}{2} \big( \| L(Y_1-Y_2)\|_F+\|(Y_1-Y_2)L_1)\|_F^2 \big) \\
\notag 
				&= \frac{1}{2} \Big[\Tr(Y_1^{\top}L^{\top}LY_1) + \Tr(LY_2 \mathbf{1}_{T_B}\mathbf{1}_{T_A}^\top)- 2 \Tr( Y_2^\top L^\top L Y_1)\\
\notag 
				&+ \Tr(Y_1L_1L_1^{\top}Y) + \Tr(Y_2^{\top} \mathbf{1}_{T_A}\mathbf{1}_{T_B}L_1L_1^\top Y_2)- 2 \Tr(Y_2 L_1L_1^{\top}Y_1^{\top} \big)\Big]
				\, .
\end{align}
We propose to use the following trick to obtain a \emph{concave} loss over $\overline{\mathcal{Y}}(X)$, the convex hull of $\mathcal{Y}(X)$. 
Let us introduce $D = \lambda_{\max}(L^{\top}L) I_{T_A \times T_A}$ and $D_1 = \lambda_{\max}(L_1L_1^{\top}) I_{T_B \times T_B}$ with $\lambda_{\max}(U)$ the largest eigenvalue of $U$. 
For any binary matrices $Y_1$, $Y_2$, we have that:
\begin{align}
\ell_S(Y_1,Y_2) &=  \frac{1}{2} \big[\Tr(Y_1^{\top}(L^{\top}L-D)Y_1) + 	
\Tr(DY_1\mathbf{1}_{T_B}\mathbf{1}_{T_A}^\top) 
\notag
\\ 
\notag 
&+ \Tr(LY_2 \mathbf{1}_{T_B}\mathbf{1}_{T_A}^\top)- 2 \Tr( Y_2^\top (L^\top L-D) Y_1)
\\ \notag
&+ \Tr(Y_1(L_1L_1^{\top}-D)Y) + 
\Tr(Y_1D_1\mathbf{1}_{T_B}\mathbf{1}_{T_A}^\top) 
\\ 
\notag &\Tr(Y_2^{\top}L_1L_1^\top Y_2)- 2 \Tr(Y_2 L_1L_1^{\top}Y_1^{\top})\Big]
\end{align}
and we get a \emph{concave} function over $\overline{\mathcal{Y}}(X)$ that coincides with $\ell_S$ on $\mathcal{Y}(X)$. 




\subsection{Empirical loss minimization}

Recall that we are given $n$ alignment examples $(X^i,Y^i)_{1 \leq i \leq n}$. For a fixed loss $\ell$, our goal is now to solve the following minimization problem in $W$:
\begin{equation}
\min_{W\in\mathcal{W}} \big\{ \frac{1}{n}\sum_{i=1}^n \ell \bigl( Y^i, \argmax_{Y\in\mathcal{Y}_{T_A^i,T_B^i}} \Tr(C(X^i;W)^\top Y) \bigr) + \lambda \Omega(W)\big\},
\label{eq:empirical}
\end{equation}
where $\Omega= \frac{\lambda}{2}\|W\|_F^2$ is a convex regularizer preventing from overfitting, with $\lambda \geq 0$. 

\section{Large margin approach}
In this section we describe a large margin approach to solve a surrogate to the problem in Eq.~\eqref{eq:empirical}, which is untractable. As shown by Eq.~\eqref{eq:decodingWithw}, the decoding task is the maximum of a linear function in the parameter $W$ and aims at predicting an output over a large and discrete space (the space of potential alignments with respect to the constraints in Eq.~\eqref{eq:constraint}). Learning $W$ thus falls in the structured prediction framework~\citep{tsochantaridis2005large,koller2003max}. 
We  define the hinge-loss, a convex surrogate to~$\ell$, by
\begin{equation}
L(X,Y;W) = \max_{Y' \in \mathcal{Y}(X)} \Big \{ \ell(Y,Y') - \Tr(W^\top \left [\phi(X,Y)-\phi(X,Y')\right ])\Big \}.
\label{eq:hinge}
\end{equation}
The evaluation of $L$ is usually referred to as the ``loss-augmented decoding'', see \citep{tsochantaridis2005large}. Among the aforementionned losses, the Hamming loss $\ell_H$ is the only one leading directly to a tractable loss-augmented decoding problem and thus that falls directly into the structured prediction framework. 
Indeed, plugging Eq.~\eqref{eq:Hammingloss} into \eqref{eq:hinge} leads to a loss-augmented decoding that is a LP over the set $\mathcal{Y}(X)$ and that can therefore be solved using the dynamic time warping algorithm. If we define $\widehat{Y}^i$ as the argmax in Eq.~\eqref{eq:hinge} when $(X,Y)=(X^i,Y^i)$, then elementary computations show that 
\BEQ
\notag 
\widehat{Y}^i = \argmin_{Y\in \mathcal{Y}_{T_A,T_B}} \tr((U^\top-2Y^{i\top}-C(X^i;W)^\top)Y),
\EEQ
where $U=\mathbf{1}_{T_B}\mathbf{1}_{T_B}^\top\in\R^{T_A\times T_B}$. 

 
We now aim at solving the following problem, sometimes called the \emph{margin-rescaled problem}:
\begin{equation}
\label{eq:structuredPredictionObjective}
\min_{W \in \mathcal{W}} \frac{\lambda}{2}\|W\|^2_F + \frac{1}{n}\sum^n_{i=1} \max_{Y \in \mathcal{Y}_{T^i_A, T^i_B}} \Big\{\ell(Y, Y^i) - \Tr(W^\top \left [\phi(X^i,Y^i)-\phi(X^i,Y)\right ]) \Big\}
\, .
\end{equation}


\paragraph{Hamming loss case.}
From Eq.~\eqref{eq:decodingWithw}, one can notice that our joint feature map is linear in $Y$. Thus, if we take a loss that is linear in the first argument of $\ell$, for instance the Hamming loss, the loss-augmented decoding is the maximization of a linear function over the spaces $\mathcal{Y}(X)$ that we can do efficiently using dynamic programming algorithms (see Sec.~\ref{sec:notations} and supplementary material).

That way, plugging the Hamming loss (Eq.~\eqref{eq:Hammingloss}) in Eq.~\eqref{eq:structuredPredictionObjective} leads to a convex structured prediction problem. This problem can be solved using standard techniques such that cutting plane methods \citep{joachims2009cutting}, stochastic gradient descent~\citep{shalev2011pegasos}, or block-coordinate Frank-Wolfe in the dual~\citep{lacoste2013block}. Note that we adapted the standard unconstrained optimization methods to our setting, where $W\succeq 0$.

\paragraph{Optimization using the symmetrized area loss.}

In this section we propose to show that it is possible to deal
The symmetrized area loss is concave in its first argument, thus the problem of Eq.~\ref{eq:structuredPredictionObjective} using it is in a min/max form and thus deriving a dual is straightforward. Details can be found in the supplementary material.
If we plug the symmetrized area loss $\ell_S$ (SAL) defined in Eq.~\eqref{eq:symloss} into our problem \eqref{eq:structuredPredictionObjective}, we can show that the dual of \eqref{eq:structuredPredictionObjective} has the following form:
\BEA \label{eq:dualObjectiveQuadraticLoss}
\min_{(Z^1, \ldots, Z^n) \in \overline{\mathcal{Y}}}& \frac{1}{2\lambda n^2} \|\sum^n_{i=1} -\sum_{j,k}(Y_i - Z^i)_{j,k}(a_j - b_k)(a_j - b_k)^T\|_F^2 \nonumber
\\&- \frac{1}{n} \sum^n_{i=1} \ell_S(Z, Z^{i}) ,
\EEA

if we denote by $\overline{\mathcal{Y}}(X^{i})$ the convex hull of the sets $\mathcal{Y}(X^{i})$, and by $\overline{\mathcal{Y}}$ the cartesian product over all the training examples $i$ of such sets. Note that we recover a similar result as \citep{lacoste2013block}. Since the SAL loss is concave, the aforrementionned problem is convex. 

The problem \eqref{eq:dualObjectiveQuadraticLoss} is a quadratic program over the compact set $\overline{\mathcal{Z}}$ . Thus we can use a Frank-Wolfe \citep{frank1956algorithm} algorithm. Note that it is similar to the one proposed by \citet{lacoste2013block} but with supplementary term due to the concavity of the loss.

\section{Experiments}\label{sec:experiments}


We applied our method to the task of learning a good similarity measure for aligning audio signals.
In this field researchers have spent a lot of efforts in designing well-suited and meaningful features \citep{joder2013learning,cont2007evaluation}. 
But the problem of combining these features for aligning temporal sequences is still challenging. 

\subsection{Dataset of \citet{kirchhoff2011evaluation}}\label{sec:experimentsChorales}

\paragraph{Dataset description.}
First, we applied our method on the dataset of \citet{kirchhoff2011evaluation}. 
In this dataset, pairs of aligned examples $(A^i, B^{i})$ are artificially created by stretching an original audio signal. That way the groundtruth alignment $Y^i$ is known and thus the data falls into our setting
A more precise description of the dataset can be found in \citep{kirchhoff2011evaluation}. 

\begin{figure}
\begin{center}
\includegraphics[scale=0.50]{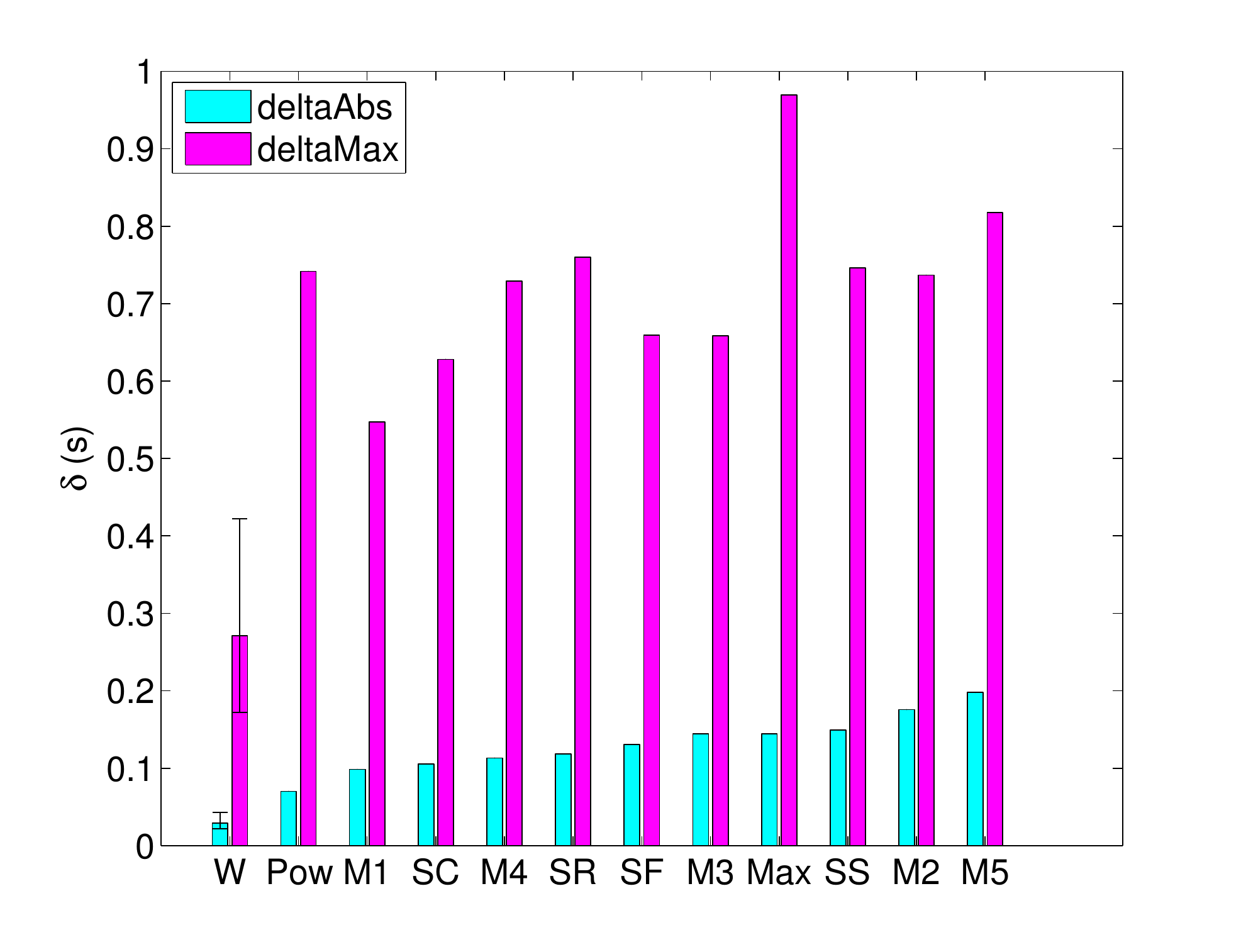}
\end{center}
\vspace*{-0.75cm}
\caption{\label{fig:feat_comp} Comparison of performance between individual features and the learned metric. The left bars are mean $\deltaabs$ error in frame on the dataset and the right are $\deltamax$. Error bars for the performance of the learned metric were determined with the best and the worst performance on $5$ different experiments.}
\end{figure}

The $N=60$ pairs are stretched along two different tempo curves. Each signal is made of $30 \second$ of music that are divided in frames of $46 \milli\second$ with a hopsize of $23\milli\second$, thus leading to a typical length of the signals of $T \approx 1300$ in our setting. 
We keep $p=11$ features simple to implement and that are known to perform well for alignment tasks \citep{kirchhoff2011evaluation}. Those were: five MFCC~\citep{gold2011speech} (labeled M1,\dots,M5 in Fig.~\ref{fig:feat_comp}), the spectral flatness (SF), the spectral centroid (SC), the spectral spread (SS), the maximum of the envelope (Max), and the power level of each frame (Pow), see~\citep{kirchhoff2011evaluation} for more details on the computation of the features. 
We normalize each feature by subtracting the median value and dividing by the standard deviation to the median, as audio data are subject to outliers. 

\paragraph{Experiments.}
We conducted the following experiment: for each individual feature, we perform alignment using dynamic time warping algorithm and evaluate the performance of this single feature in terms of losses typically used to asses performance in this setting \citep{kirchhoff2011evaluation}. In Fig.~\ref{fig:feat_comp}, we report the results of these experiments.

Then, we plug these data into our method, using the Hamming loss to learn a linear positive combination of these features. The result is reported in Fig~\ref{fig:feat_comp}. Thus, combining these features on this dataset yields to better performances than only considering a single feature. 

\subsection{Chorales dataset}

\paragraph{Dataset.}
The Bach $10$ dataset\footnote{\url{http://music.cs.northwestern.edu/data/Bach10.html}.} consists in ten J. S. Bach's Chorales (small quadriphonic pieces). For each Chorale, a MIDI reference file corresponding to the ``score'', or basically a representation of the partition. The alignments between the MIDI files and the audio file are given, thus we have converted these MIDI files into audio following what is classically done for alignment (see e.g, \citep{hu2003polyphonic}). That way we fall into the audio-to-audio framework in which our technique apply. Each piece of music is approximately $25\second$ long, leading to similar signal length. 

\begin{figure}
\begin{center}
\includegraphics[scale=.33]{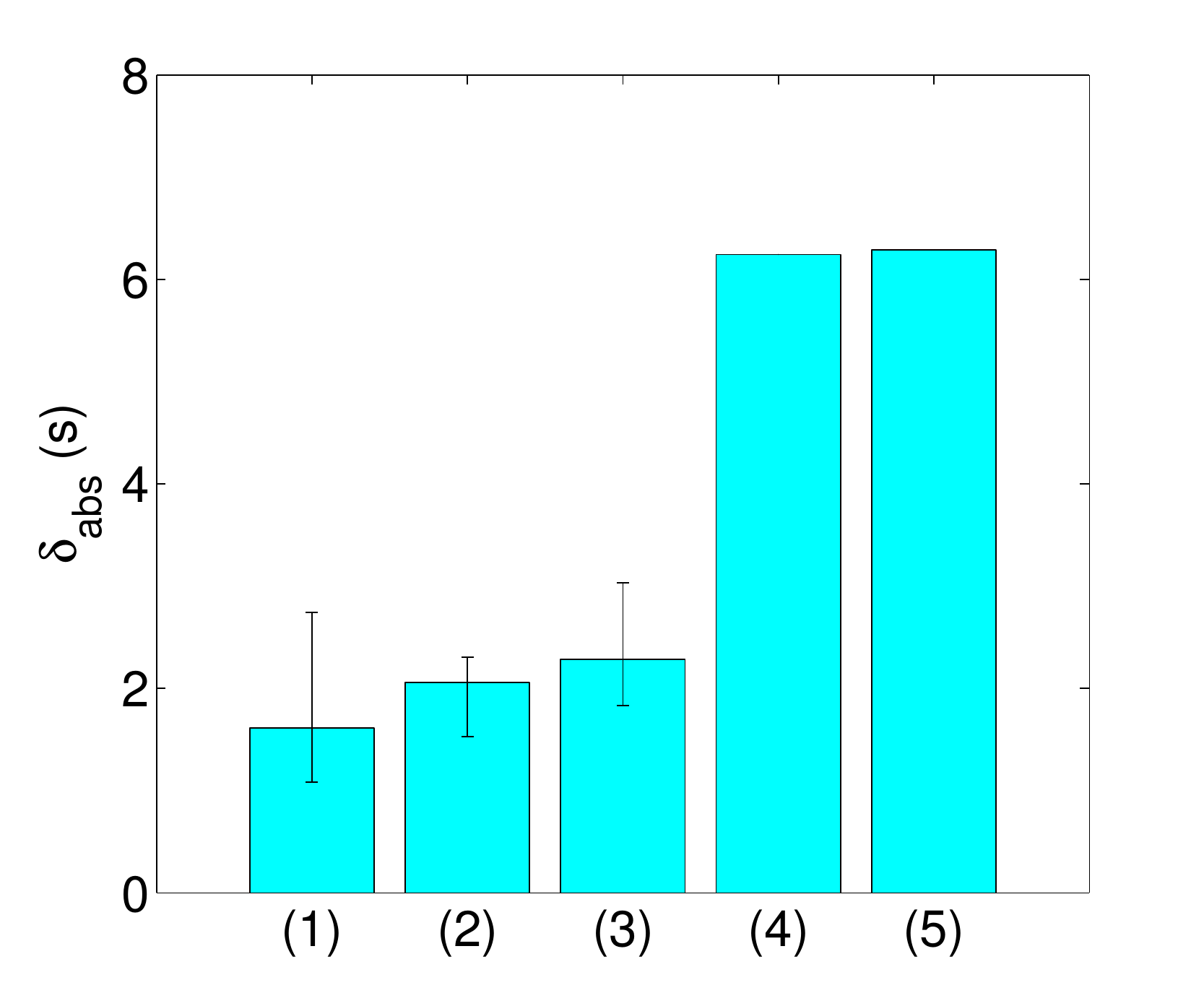}
\end{center}
\vspace*{-0.5cm}
\caption{\label{fig:perf_comp} Performance of our algorithms on the Chorales dataset. From left to right: (1) Best single feature, (2) Best learned combination of features using the symmetrized area loss $\ell_S$, (3) Best combination of MFCC and derivatives learned with $\ell_S$, (4) Best combination of MFCCs and derivatives learned with Hamming loss, (5) Best combination of features of \citep{kirchhoff2011evaluation} using Hamming loss.}
\end{figure}

\paragraph{Experiments.}
We use the same features as in Sec.~\ref{sec:experimentsChorales}. As depicted in Fig.~\ref{fig:perf_comp}, the optimization with Hamming loss performs poorly on this dataset. In fact, the best individual feature performance is far better than the performance of the learned $W$. Thus metric learning with the ``practical'' Hamming loss performs much worse than the best single feature. 

Then, we conducted the same learning experiment with the symetrized area loss
 $\ell_S$. The resulting learned parameter is far better than the one learned using the Hamming loss. We get a performance that is similar to the one of the best feature. Note that these features were handcrafted and reaching their performance on this hard task with only a few training instances is already challenging. 

In Fig.~\ref{fig:hamming_fail}, we have depicted the result, for a learned parameter $W$, of the loss augmented decoding performed either using the area. As it is known for structured SVM, this represents the most violated  constraint \cite{tsochantaridis2005large}. We can see that the most violated constraint for the Hamming loss leads to an alignment which is totally unrelated to the groundtruth alignment whereas the one for the symmetrized area loss is far closer and much more discriminative.

\subsection{Feature selection}

Last, we conducted experiments over the same datasets. Starting from low level features, namely the $13$ leading MFCCs coefficients and their first two derivatives, we learn a linear combination of these that achieves good alignment performance in terms of the area loss.Note that very little musical prior knowledge is put into these. Moreover we either improve on the best handcrafted feature on the dataset of \citep{kirchhoff2011evaluation} or perform similarly. On both datasets, the performance of learned combination of handcrafted features performed similarly to the combination of these $39$ MFCCs coefficients

\section{Conclusion}

In this paper, we have presented a structured prediction framework for learning the metric in temporal alignment problems. We were able to combine hand-crafted features, as well as building automatically new state-of-the-art features from basic low-level information with very little expert knowledge.

Technically, this is made possible by considering a loss beyond the usual Hamming loss which is typically used because it is ``practical'' within a structured prediction framework (linear in the output representation).

The present work may be extended in several ways, the main one being to consider cases where only partial information about the alignments is available. This is often the case in music~\citep{cont2007evaluation} or bioinformatics applications. Note a simple alternating optimization between metric learning and constrained alignment provide a simple first solution, which could probably be improved upon.

\subsection*{Acknowledgements}
We acknowledge the support of the GARGANTUA project
(Mastodons program of CNRS), the grant SIERRA-23999
from the European Research Council and a PhD fellowship
from the EADS Foundation.

{
\bibliographystyle{abbrvnat}
\bibliography{timewarping}
}

\appendix

\section{Derivation of the BCFW-like algorithm for the quadratic loss.}

\subsection{Relaxing the set for loss augmented inference.}

Let us start from the global structured objective equation of the paper. Recall that we dispose of the training examples $((X^1,Y^1),\dots,(X^n,Y^n))$. In order to make the derivation easier, and following \citet{lacoste2013block}, we denote the difference between the feature map associated to any $Y \in \mathcal{Y}(X^i)$ and the one associated to the true training example label $Y_i$ by:  $ \Tr(W\phi(X^{i}, Y^{i})) = \Tr(W\sum_{j,k}(Y_{j,k}^i-Y_{j,k})(a^i_j-b_k^i)(a^i_j-b_k^i)^T) = \langle W, \psi^i(Y) \rangle$. The objective of structured prediction is thus:
\BEQ \label{eq:objectiveNonRelaxedSet}
\min_{W \in \mathcal{W}} \frac{\lambda}{2}\|W\|^2_2 + \frac{1}{n}\sum^n_{i=1} \max_{Y \in \mathcal{Y}(X^i)} \left\{\ell_i(Y, Y^i) - \langle W, \psi^i(Y) \rangle \right\}.
\EEQ
The term $\max_{Y \in \mathcal{Y}(X^i)} \left\{\ell_i(Y, Y^i) - \langle W, \psi^i(Y) \rangle \right\}$ corresponds to the structural hinge loss for our problem.
Let us introduce $\overline{\mathcal{Y}}(X^i)$ the convex hull of the sets $\mathcal{Y}(X^i)$. We will also use $\overline{\mathcal{Y}} = \overline{\mathcal{Y}}(X^1)\times \ldots \times \overline{\mathcal{Y}}(X^n)$. From now on, we will perform the loss augmented decoding on this relaxed set. This problem has potentially \emph{non integral} solutions. We call the maximization of the hinge loss over $\overline{\mathcal{Y}}$ the \emph{loss augmented inference}.
Now we can write a new optimization objective:
\BEQ \label{eq:objectiveRelaxedSet}
\min_{W \in \mathcal{W}} \frac{\lambda}{2}\|W\|^2_2 + \max_{(Z_1, \ldots, Z_n) \in \overline{\mathcal{Y}}} \left\{\frac{1}{n}\sum^n_{i=1} \big [ \ell_i(Z_i, Y^i) - \langle W, \psi^i(Z_i) \rangle \big ] \right \}.
\EEQ
Note that since our joint feature map $\phi(X^{i}, Y)$ is linear in $Y$, if $\ell$ is linear as well (for instance if $\ell$ is the Hamming loss), this problem is strictly equivalent to~\eqref{eq:objectiveNonRelaxedSet} since  in that case, the loss-augmented inference is a LP over $\overline{\mathcal{Y}}(X^i)$, which has necessary a solution in $\mathcal{Y}(X^i)$ (see, e.g, [Prop. B.21] of \citep{bertsekas1999nonlinear}.

In general, in order to be convex and thus tractable, the aforrementioned problem requires a loss which is concave over the convex sets $\overline{\mathcal{Y}}(X^i)$.

\subsection{Dual of the structured SVM}

Since Prob.~\eqref{eq:objectiveNonRelaxedSet} is in saddle point form, we get the dual by switching the max and the min:
\BEQ
\max_{(Z_1, \ldots, Z_n) \in \overline{\mathcal{Y}}(X^1)\times \ldots \times \overline{\mathcal{Y}}(X^n)} \min_{W \in \mathcal{W}} \frac{\lambda}{2}\|W\|^2_2 + \{\frac{1}{n}\sum^n_{i=1} \big [ \ell_i(Y, Y^i) - \langle W, \psi^i(Z_i) \rangle \big ] \}.
\EEQ
From the above equation, we deduce the following relation linking primal variable $W$ and dual variables $(Z_1, \ldots, Z_n) \in \mathcal{Y}(X^1)\times \ldots \times \mathcal{Y}(X^n)$:

In the specific case when $\mathcal{W}$ is unconstrained and simply equals to $\mathbb{R}^{p \times p}$, this reduces to:
\BEQ \label{eq:representerTheorem}
W = \frac{1}{\lambda} \sum^n_{i=1}\psi_i(Z_i).
\EEQ
If $\mathcal{W}$ is the set of symmetric semidefinite positive matrices we get:
\BEQ \label{eq:representerTheoremSDP}
W = \frac{1}{\lambda} \sum^n_{i=1}(\psi_i(Z_i))_+,
\EEQ
with $(\psi_i(Z_i))_+$ the projection of $(\psi_i(Z_i))$ over $\mathcal{W}$.

Eventually if we consider $\mathcal{W}$ the set of diagonal matrices, and denote by Diag the operator associating to a matrix the matrix composed of its diagonal:
\BEQ \label{eq:representerTheoremDiag}
W = \frac{1}{\lambda} \sum^n_{i=1}\rm{Diag}(\psi_i(Z_i)).
\EEQ
These relations are also known as the ``representer theorems''.

For what follows we consider the case of $\mathcal{W}=\mathbb{R}^{p \times p}$ but dealing with the other cases is similar.

In that case the dual can be written simply as:
\BEQ \label{eq:dualProblem1}
\max_{(Z_1, \ldots, Z_n) \in \overline{\mathcal{Y}}(X^1)\times \ldots \times \overline{\mathcal{Y}}(X^n)} -\frac{1}{2\lambda n^2} \|\sum^n_{i=1} \psi_i(Z_i)\|_F^2 + \frac{1}{n}\sum^n_{i=1} \ell(Y_i, Z_i).
\EEQ
We recover a result similar to the ones of~\citet{lacoste2013block}.

\subsection{A Frank-Wolfe algorithm for solving Prob.~\eqref{eq:dualProblem1}}

Now, we can derivate a Frank-Wolfe algorithm for solving the dual problem of \ref{eq:dualProblem1}.
As noted in the paper, we are able to maximize or minimize any linear form over the sets $\mathcal{Y}(X^i)$, thus we are able to solve LPs over the convex hulls $\overline{\mathcal{Y}}(X^i)$ of such sets.

Plugging back the specific form of our joint feature map directly into Eq.~\eqref{eq:dualProblem1} we get that $\psi_i(Z^i) = -\sum_{j,k}(Y_i - Z^i)_{j,k}(a_j - b_k)(a_j - b_k)^T$ and thus we can write the dual problem as:
\BEA
\label{eq:dualProblem2}
\min_{\substack{(Z^1, \ldots, Z^n) \in \\ \overline{\mathcal{Y}}(X^1)\times \ldots \times \overline{\mathcal{Y}}(X^n)}}
\frac{1}{2\lambda n^2} \|\sum^n_{i=1} -\sum_{j,k}(Y_i - Z^i)_{j,k}(a_j - b_k)(a_j - b_k)^T \|_F^2 - \frac{1}{n}\sum^n_{i=1} \ell(Y^i, Z^i)
\EEA
Now, as in the paper, let us introduce $L \in \mathbb{R}^{T_A \times T_A}$ and $L_1 \in \mathbb{R}^{T_B \times T_B}$. If $U_i$ is the matrix of ones of the same size as $Z^i$, we consider the following loss: 
\BEA
\ell(Y^i, Z^i) &=& \frac{1}{2}\big [\Tr(Z^{iT}(L^TL-D)Z^i) + \Tr(DZ^iU^i) + \Tr(Y^{iT}L^TL^i) - 2 \Tr(Z^{iT}L^TLY^i) \nonumber \\ &+& \Tr(Z^{i}(L_1^TL_1-D_1)Z^i)+ \Tr(D_1Z^iU^i) + \Tr(Y^{i}L_1^TL_1^i) - 2 \Tr(ZL_1^TL_1Y^i) \big ].
\EEA
This loss is sound for alignments problems since, when $Y_i$ and $Z^i$ are in $\mathcal{Y}$, this is simply the $\ell_S$ loss $\|LY_i - LZ^i\|_F^2 + \|Y_iL_1 - Z^iL_1\|_F^2$.

Thus we get the following overall dual objective:
%
\BEA \label{eq:dualObjectiveQuadraticLossapp}
\min_{(Z^1, \ldots, Z^n) \in \overline{\mathcal{Y}}} & &
\frac{1}{2\lambda n^2} \|\sum^n_{i=1} -\sum_{j,k}(Y_i - Z^i)_{j,k}(a_j - b_k)(a_j - b_k)^T\|_F^2 \nonumber \\ 
&-& \frac{1}{n} \big(\sum^n_{i=1} [\Tr(Z^{iT}(L^TL-D)Z^i) + \Tr(Z^{iT}DU^i) + \nonumber \\ & &\Tr(Y^{T}L^TL^i) - 2 \Tr(Z^{iT}L^TLY^{i}) + \Tr(Z^{i}(L_1^TL_1-D_1)Z^{iT}) \nonumber \\
&+& \Tr(U_iD_1Z^i) + \Tr(YL_1^TL_1^i) - 2 \Tr(ZL_1^TL_1Y^{iT}) \big ]).
\EEA
We recall that $D$ is a diagonal matrix such that $A^\top A-D \preceq 0$ and thus our objective is convex.
Our dynamic programming algorithm (DTW) is able to maximize any linear function over the sets. Thus we can use a Frank-Wolfe \citep{frank1956algorithm} technique.
At iteration $t$, this algorithm iteratively computes a linearization of the function at the current point $(Z^{1},\ldots Z^{n})_k$, computes a linearization of the function, optimize it, get a new point $(Z^{1},\ldots Z^{n})^{*}_k$and then make a convex combination using a stepsize $\gamma$.
 
Note that we have directly a stochastic version of such an algorithm. As noted in \citet{lacoste2013block} instead of computing a gradient for each block of variable $Z^{i}$, we simply need to choose randomly one block at each timestep and make an update on these variables.

The linearization simply consists in computing the matrix gradient for each of the matrix variables $Z^i$ which turns out to be:
\BEA
\nabla_{Z^i}(g) = & \frac{1}{n} \big[  C - \frac{1}{2} \big( 2(L^{T}L - D)Z^{i} + DU_i - 2L^{T}LY^{i} \nonumber \\ +& 2Z^{i}(L_1^{T}L_1 - D) + U_iD - 2Y^{i}L^{T}L\big)\big]  
\EEA

where $C$ is simply the affinity matrix of dynamic time warping.

\section{The dynamic time warping algorithm}

Let us give the pseudocode of the dynamic time warping that maximize the LP~(2) of the article.
In opposition to~\citet{muller2007information}, we give a version of the algorithm for the affinity matrix $C$. Intuitively, the cost matrix is the opposite of a cost matrix, thus we aim to maximize the cumulated affinity instead of minimizing the cumulated cost.
This algorithm is $O(T_AT_B)$, making it very costly to compute for large time series.
 
\begin{figure}[h]
\begin{algorithmic}
\STATE{Cumulated affinity matrix:}
\STATE{$T,S\leftarrow \mathrm{size}(C)$, $D\leftarrow \mathrm{zeros}(T_A+1,T_B+1)$}
\FOR{$i=1$ \TO $T_A$}
\STATE{$D(i,0)\leftarrow -\infty$}	
\ENDFOR
\FOR{$j=1$ \TO $T_B$}
\STATE{$D(0,j)\leftarrow -\infty$}	
\ENDFOR
\FOR{$i=1$ \TO $T_A$}
\FOR{$j=1$ \TO $T_B$}
\STATE{$D(i,j)\leftarrow C(i,j) + \max(D(i-1,j),D(i,j-1),D(i-1,j-1))$}
\ENDFOR
\ENDFOR
\STATE{Backtracking:}
\STATE{$Y\leftarrow \mathrm{zeros}(T_A,T_B)$, $i\leftarrow T_A$, $j\leftarrow T_B$}
\WHILE{$i>1$ \OR $j>1$}
\STATE{$Y(i,j)\leftarrow 1$}
\IF{$i == 1$}
\STATE{$j\leftarrow j-1$}
\ELSIF{$j == 1$}
\STATE{$i \leftarrow i-1$}
\ELSE
\STATE{$m\leftarrow\max(D(i-1,j),D(i,j-1),D(i-1,j-1))$}
\IF{$D(i-1,j) == m$}
\STATE{$i\leftarrow i-1$}
\ELSIF{$D(i,j-1) == m$}
\STATE{$j\leftarrow j-1$}
\ELSE
\STATE{$i\leftarrow i-1$, $j\leftarrow j-1$}
\ENDIF
\ENDIF
\ENDWHILE
\RETURN{Y}
\end{algorithmic}
\caption{The dynamic time-warping algorithm that solves the LP~(2), for a given similarity matrix $C$.\label{algo:dtw}
}
\end{figure}

\end{document}